\documentclass[10pt,journal,final,finalsubmission,twocolumn]{IEEEtran}

\usepackage{cite}
\usepackage{graphicx}
\usepackage{psfrag}
\usepackage{subfigure}
\usepackage{url}
\usepackage{amsmath}
\usepackage[ruled,vlined]{algorithm2e}
\usepackage{multirow}
\usepackage{multicol}
\usepackage{algorithmic}
\usepackage{array}
\usepackage{mdwmath}
\usepackage{mdwtab}
\usepackage{eqparbox}
\usepackage{amsmath}
\usepackage{amssymb}
\usepackage{amsfonts}
\usepackage{epsfig}
\usepackage{color}

\usepackage{array}

\begin{document}
\title{A Tiered Move-making Algorithm for General Non-submodular Pairwise Energies}
\author{Vibhav~Vineet,
        Jonathan~Warrell,
        and~Philip H.S.~Torr
\thanks{email: vibhav.vineet@gmail.com}
\thanks{}}

\maketitle

\begin{abstract}

A large number of problems in computer vision can be modelled as energy minimization problems in a Markov Random Field (MRF) or Conditional Random Field (CRF) framework. Graph-cuts based $\alpha$-expansion is a standard move-making method to minimize the energy functions with sub-modular pairwise terms. However, certain problems require more complex pairwise terms where the $\alpha$-expansion method is generally not applicable. 

In this paper, we propose an iterative {\em tiered move making algorithm} which is able to handle general pairwise terms. Each move to the next configuration is based on the current labeling and an optimal tiered move, where each tiered move requires one application of the dynamic programming based tiered labeling method introduced in Felzenszwalb et. al. 
\cite{tiered_cvpr_felzenszwalbV10}. The algorithm converges to a local minimum for any general pairwise potential, and we give a theoretical analysis of the properties of the algorithm, characterizing the situations in which we can expect good performance. We first evaluate our method on an object-class segmentation problem using the Pascal VOC-11 segmentation dataset where we learn general pairwise terms. Further we evaluate the algorithm on many other benchmark labeling problems such as stereo, image segmentation, image stitching and image denoising. Our method consistently gets better accuracy and energy values than $\alpha$-expansion, loopy belief propagation (LBP), quadratic pseudo-boolean optimization (QPBO), and is competitive with TRWS.
\end{abstract}

\begin{keywords}
Energy Minimization, Markov Random Fields, Move making algorithm, Tiered labebling method, Dynamic programming
\end{keywords}

\section{Introduction}

A large number of problems in computer vision can be modeled as discrete labeling problems.  Examples include depth estimation from stereo, object segmentation and scene understanding, photomontage and image denoising \cite{Boykov_pami01,tiered_cvpr_felzenszwalbV10,kohli_p3_pami09,kolmogorov_qpbo}.  Random fields provide a general framework
for building models for such problems, and can capture a wide range of phenomena which are relevant in many cases, such as smoothness and local interactions between labels.
\begin{figure}[t]
\begin{center}
  \begin{tabular}{cc}
      \includegraphics[width=0.45\linewidth]{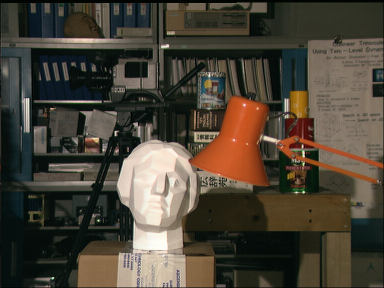}
      \includegraphics[width=0.450\linewidth]{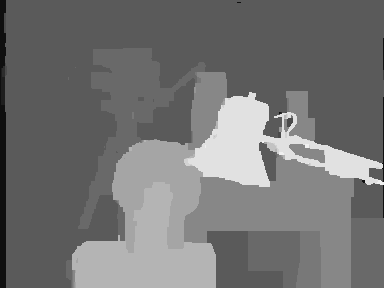}
  \end{tabular}
\end{center}
\caption{\textit{Our tiered move making algorithm achieves a minimum energy value $0.3\%$ lower that that achieved by $\alpha$-expansion with a truncated L1-norm pairwise potential for
 stereo on the tsukuba image from \cite{Szeliski_pami08} with 16 disparity labels. Further, we achieve around $0.01\%$, $0.1\%$ and $1\%$ lower energy values with Potts, 
 non-truncated L1 and L2-norm pairwise potentials respectively. See experimental section for more details.}}
\label{fig:stereo_images}
\end{figure}

In general, such problems are NP-hard.  Certain subclasses though are known to have polynomial time algorithms.  Examples of these include binary labeling problems with submodular pairwise interaction terms \cite{kolmogorov_qpbo}, and multiple label problems with convex pairwise interactions \cite{ishikawa}.  The former can include for example foreground/background segmentation problems, while the latter can naturally be applied to cases in which an ordering exists across the labels, as in stereo. In both cases, the problem can be solved by building a graph to represent the problem, and solving an {\em s-t min cut} problem on the graph. Constraining all problems to take such forms however is restrictive, and can lead to poor models in general.  Examples of problems that cannot naturally be cast in such ways include multi-label problems without a label ordering (e.g. object class segmentation \cite{kohli_p3_pami09}), problems with an ordered label set but a truncated pairwise cost (e.g. state-of-the-art stereo models \cite{kolmogorov_qpbo}), and binary problems with non-submodular pairwise terms. Further, in some of the problem domains such as object-class segmentation and texture restoration, it can be beneficial to learn a function for the pairwise  terms on training data, which will not in general respect such constraints. Learning general pairwise terms captures the true prior on the dataset, and has proved to improve the overall accuracy \cite{krahenbuhl_nips11},\cite{kolmogorov_qpbo}.

A number of methods have been introduced which can be applied to solve these problems approximately. Approaches include belief propagation \cite{freeman_gbp_nips00}, linear programming relaxations \cite{Kolmogorov_trws_pami06,komodakis2007}, dynamic programming based approaches \cite{veksler_fdp_cvpr12}, \cite{veksler_fdp_eccv12} \cite{tiered_cvpr_felzenszwalbV10} and move-making algorithms \cite{Boykov_pami01,gouldAlphabetSoup,kumar2011improved,lempitsky2010fusion}. We will focus on the last of these approaches, although we note that connections have recently been drawn between graph-cut based $\alpha$-expansion moves and linear programming \cite{komodakisGraphCuts}, as well as graph-cuts and belief propagation \cite{linkGraphCutsMP}, pointing to a fundamental unity of these methods.

Move-making algorithms are designed to handle multi-label problems.  They proceed by maintaining a {\em current solution} to the problem, and updating this in a series of moves, in which all or a subset of the pixels are allowed either to maintain their current label, or switch to a possibly restricted subset of labels.  Moves are made only if they decrease the global energy of the random field, and termination occurs when no further move can decrease the energy \cite{gouldAlphabetSoup}.  Key factors which determine the effectiveness of a move-making algorithm are the suitability of the {\em move space} for the problem at hand, and the guarantees that can be made of the algorithm used to search this space.  The $\alpha$-expansion algorithm \cite{Boykov_pami01} for instance is typically effective on problems where large contiguous regions sharing the same label can be expected in the solution (e.g. segmentation problems), as the move space allows groups of pixels to switch to a common label in one move, and the optimal move can be found at each step for metric pairwise terms, which help to enforce piecewise smoothness.  In other cases, acceptance of suboptimal moves can be compensated for by especially designed move-spaces.  For instance, high-quality independently generated solutions can be exploited when available to propose alternative labels at each pixel as in {\em fusion moves} \cite{lempitsky2010fusion,woodford2006fields}, where QPBO can be used as the move generator \cite{kolmogorov_qpbo}. Alternatively, the fact that large regions of the solution are expected to use restricted ranges of labels from an ordered set (e.g. stereo) can be exploited by using moves which depend on a range, allowing each pixel to swap to a label from that range as in {\em range expansion moves} \cite{kumar2011improved}, where the move generator is derived from 
\cite{ishikawa}.

From the approaches discussed above, while \cite{Boykov_pami01} uses exact (optimal) moves, it places metric conditions on the pairwise terms, and while \cite{kumar2011improved,lempitsky2010fusion} can be applied to more general energies, \cite{lempitsky2010fusion} relies on independently generated solutions, \cite{kumar2011improved} can only be applied to energies with truncated convex pairwise terms, and neither uses optimal moves. The question arises then whether a move-making approach can be proposed using an exact algorithm to generate optimal moves for general multi-label energies.\footnote{We note that \cite{gouldAlphabetSoup} analyze $\gamma$-expansion moves, which are optimal in a stricter sense than we consider here.  Our tiered moves make an optimal search over a fixed large space of transformations (Sec. \ref{sec:move}).}  Here, we propose such an algorithm based on the dynamic programming approach to tiered labeling problems introduced in \cite{tiered_cvpr_felzenszwalbV10}. Our move space is thus defined by the tiered labeling constraints: at each move, a band of pixels across the image may change their labels, and each column of the image may choose a different label to change to. However, since multiple moves are made, the overall solution can be an arbitrary labeling. As we demonstrate, such moves are effective for a wide range of problems, for instance in which large contiguous label regions with convex shape are expected (segmentation of certain object classes) or many piecewise constant vertical regions are expected (stereo, see Fig.~\ref{fig:stereo_images}). We also show that the algorithm remains effective on problems where these assumptions do not obviously hold (photomontage, and image denoising). Further, to demonstrate the generalization ability of our algorithm in handling arbitrary pairwise terms, we evaluate it on an object-class segmentation problem using the Pascal VOC-11 segmentation dataset where pairwise terms are learnt on the training set. We observe an improvement in the overall accuracy from using learnt rather than predefined pairwise terms, where graph-cuts based $\alpha$-expansion cannot be used with learnt terms because of non-submodularity. We compare against a range of other approximate algorithms, including $\alpha$-expansion, loopy belief propagation (LBP), quadratic pseudo-boolean optimization (QPBO) and sequential tree reweighted message passing (TRWS), and show our algorithm consistently to achieve lower energies than all except for TRWS, with which we remain competitive.  Finally, the fact we use optimal moves allows us to guarantee a bound on the quality of the global solution for certain classes of energy, which we show (under certain conditions) reduces to the $\alpha$-expansion bound when a metric pairwise term is used. A previous version of this paper appeared as \cite{vineetwt_cvpr12}. This extended version contains experiments exploring the effects of performing tiered moves in both horizontal and vertical directions, and an extended analysis of time complexity of our method. It also contains experiments on the Pascal VOC-11 segmentation dataset. 

In Sec. \ref{sec:tiered} we review the tiered labeling algorithm of \cite{tiered_cvpr_felzenszwalbV10}.  We then show how this can be used as the basis of a move making algorithm on arbitrary energies in Sec. \ref{sec:move}.  Sec. \ref{sec:bound} provides a theoretical analysis of the algorithm, characterizing its optimality and complexity properties, Sec. \ref{sec:exp} gives an experimental comparison with other approaches, and Sec. \ref{sec:disc} concludes with a discussion.
\vspace{-1mm}
\subsection{Tiered Labeling}\label{sec:tiered}
\vspace{-1mm}
We review here Felzenszwalb et al's {\em tiered labeling} algorithm \cite{tiered_cvpr_felzenszwalbV10}, which we will use to form the basis of our move-making algorithm. The tiered labeling algorithm performs MAP inference in a pairwise MRF/CRF with general forms for the potentials subject to certain constraints on the final label configurations.

We let $p\in P$ range over the pixels of an image, and let $\mathbf{f} \in \hat{L}^{|P|}$ represent a labeling of the image, where $\hat{L}$ is a discrete label set, and we write $f_p=l$ to denote that pixel $p$ is assigned label $l$.  A pairwise MRF/CRF can be defined by an energy model, $E:f\rightarrow R$:
\begin{eqnarray}\label{eq:mrf}
E(f) = \sum_{p\in P}D_p(f_p) + \sum_{pq\in N} V_{pq}(f_p, f_q)
\end{eqnarray}
where $D_p(f_p)$ and $V_{pq}(f_p, f_q)$ are the unary and pairwise term respectively, and $N$ is the set of neighbors for each pixel $p$ in the image.  For the tiered labeling algorithm, we assume there are two special members of $\hat{L}$, which we label $T$ and $B$, for top and bottom.  Writing $r_p\in\{1...m\}$ and $k_p\in\{1...n\}$ respectively for the row and column of pixel $p$ in the image, where $m$ and $n$ are the number of rows and columns, we can express the constraints on the allowable tiered 
labeling solutions as:
\begin{equation}
\forall k \leq m \; \exists i, j \leq n, l \in \hat{L}\backslash\{T,B\} \; \text{s.t.} \nonumber
\end{equation}
\begin{eqnarray}\label{eq:constraints}
k_p = k \Rightarrow ( ( r_p < i \Rightarrow f_p = T ) \wedge (r_p \geq j \Rightarrow f_p = B ) \nonumber \\
\wedge (i \leq r_p < j \Rightarrow f_p = l))
\end{eqnarray}
That is, we are allowed only two breakpoints $i$ and $j$ on each column, such that above $i$ the column is labeled $T$, below (and including) $j$ it is labeled $B$, and between the two it takes a single label $l$ from the remaining labels (where $l$ can vary with the column).  When $i=j$ label $l$ is not used, and only $T$ and $B$ appear on the column.

Felzenszwalb et al. are able to minimize Eq. \ref{eq:mrf} subject to Eq. \ref{eq:constraints} exactly using dynamic programming.  To do this, they first reexpress 
Eq. \ref{eq:mrf} as:
\begin{eqnarray}\label{eq:dp}
E(f) = \sum_{k=1}^{n}U_k{(s_k)} + \sum_{k}^{n-1}H_k{(s_k, s_{k+1})}
\end{eqnarray}
Here, $k$ ranges over the $n$ columns of the image.  Instead of expressing the energy in terms of pixel label assignments, $f_p$, Eq. \ref{eq:dp} is expressed in terms of the variables $s_{k=1...n}$.  These variables have the structure $s_k = (i_k,\;j_k,\;l_k)$, where $i_k$ and $j_k$ represent the two break-points for column $k$, as described above, and $l_k\in \hat{L} \backslash \{T,B\}$ is the label to be assigned to the pixels of the column for which $i_k \leq r_p < j_k$.  Any collection 
$\{s_1, s_2, ... s_{n}\}$ thus defines a labeling $\mathbf{f}$.  The terms $U$ and $H$ represent unary and pairwise terms in this reformulated energy, and can be defined in terms of the previous energy terms by letting $P_k = \{p|k_p=k\}$, $N_k = \{(p,q)|(p,q)\in N, k_p=k_q=k\}$, $N_{k_1,k_2} = \{(p,q)|(p,q)\in N, k_p=k_1, k_q=k_2\}$, and 
$f_p(s_k)$ be the labeling of pixel $p\in P_k$ implied by $s_k$ as:
\begin{eqnarray}\label{eq:dp1}
&& U_k(s_k) = \sum_{p\in P_k}D_p(f_p(s_k))+\sum_{pq\in N_k} V_{pq}(f_p(s_k), f_q(s_k)) \nonumber \\
&& H_k(s_k,s_{k+1}) = \sum_{pq\in N_{k,k+1}} V_{pq}(f_p(s_k), f_q(s_{k+1})) 
\end{eqnarray}
Given these definitions, it is clear that minimizing Eq. \ref{eq:dp} with respect to $s_{1...n}$ is equivalent to minimizing Eq. \ref{eq:mrf} with respect to $f$, subject to the constraints in Eq. \ref{eq:constraints}.

Since Eq. \ref{eq:dp} defines an energy on a chain, it can be solved via dynamic programming (DP).  Writing $n$ and $m$ for the number of columns and rows respectively, and $K$ for the number of labels (disregarding $T$ and $B$), a naive DP implementation would have time complexity $O(nm^4K^2)$, since the number of states for variable $s_k$ is $m^2K$.  However, \cite{tiered_cvpr_felzenszwalbV10} introduce two tricks to reduce the time complexity to $O(nm^2K^2)$ by utilizing the special grid structure of MRF. The first involves the search for the best $(i',j')$ in column $k-1$ given $(i,j)$ in the column $k$, which would require $O(m^2)$ complexity for a brute-force search, but can be achieved in $O(m)$ time through a decoupling strategy.  Second, the calculation of $U_k(s_k)$ and $H_k(s_k, s_{k-1})$ can be computed in $O(1)$ time using cumulative 
sums of the $U$, and $H$ respectively. Further details on how the reduction of complexity is achieved can be found in \cite{tiered_cvpr_felzenszwalbV10}.

\begin{figure}[t]
\begin{center}
  \begin{tabular}{cc}
     \includegraphics[width=0.30\linewidth]{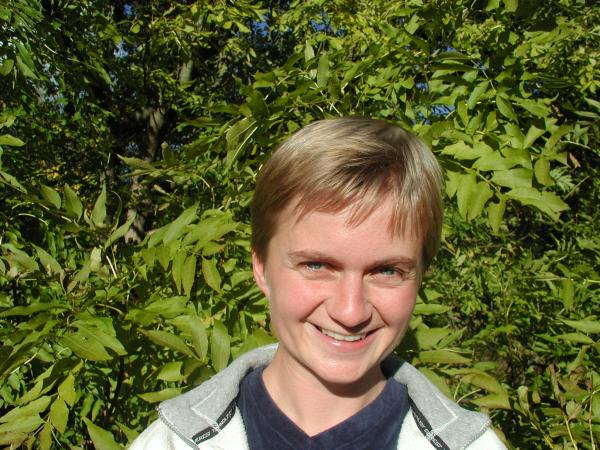}
     \includegraphics[width=0.30\linewidth]{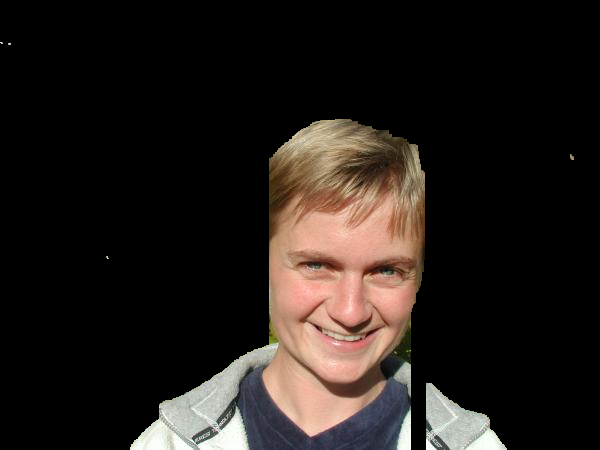}
     \includegraphics[width=0.30\linewidth]{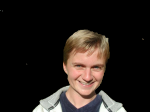} \\
     \includegraphics[width=0.30\linewidth]{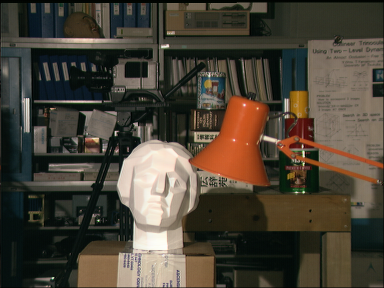}  
     \includegraphics[width=0.30\linewidth]{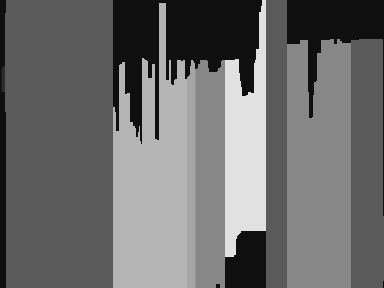} 
     \includegraphics[width=0.30\linewidth]{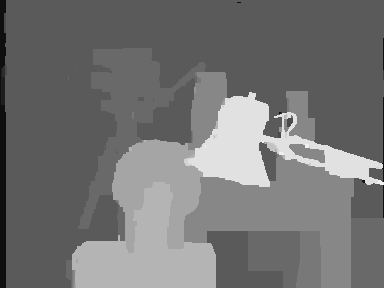}     
  \end{tabular}
\end{center}
\caption{\textit{Shown are some of the failure cases of the tiered labeling method~\cite{tiered_cvpr_felzenszwalbV10} on scenes which do not meet tiered constraints. Here we show that our method is able to achieve better results (third column) than the tiered labeling method (second column) on segmentation (first row) and stereo correspondence problems (second row).}}
\label{fig:tiered_success_fail}
\end{figure}

Tiered labeling has been successful in achieving globally optimal solutions on certain geometric structures such as road scenes, though it performs poorly on other general shapes and labelling problems. Fig.~\ref{fig:tiered_success_fail} shows some failure cases of the tiered labeling method.
\vspace{-1mm}
\section{Tiered move-making algorithm}\label{sec:move}
In general, a move making algorithm can be expressed in terms of a {\em space of moves}, $t \in \mathcal{T}$, and a {\em transformation function} $\mathbb{T}$, which takes as input a labeling $f^1$ and transformation $t$, and outputs a second labeling $f^2$ (i.e. the result of applying $t$ to $f^1$): $\mathbb{T}(f^1,t)=f^2$.  The algorithm proceeds by generating a series of labelings across iterations $\nu = 1, 2, ... I$, such that either $E(f^{\nu+1}) = \arg\min_{t\in\mathcal{T}} E(\mathbb{T}(f^{\nu},t))$, or $E(f^{\nu+1}) = \arg\min_{t\in\mathcal{T}^\prime} E(\mathbb{T}(f^{\nu},t))$, where $\mathcal{T}^\prime \subset \mathcal{T}$.  For some move spaces, the minimization cannot be made exactly, in which case a move $t$ is chosen such that $E(f^{\nu+1}) = E(\mathbb{T}(f^{\nu},t)) \leq E(f^{\nu})$.  The algorithm terminates when 
$E(f^{\nu+1}) \geq E(f^{\nu})$.

For our tiered move-making algorithm, we shall assume that we are minimizing a discrete energy function of the form Eq. \ref{eq:mrf}, where $f_p = l \in L$ is a member of a generic label set $L$ with no special members (such as $T$ and $B$ in Sec. \ref{sec:tiered}) and subject to no constraints. We then consider the following move space, expressed in terms of vectors $\mathbf{t}$ over the augmented space of labels $\hat{L} = L\cup\{T,B\}$, i.e. $\mathbf{t} \in \hat{L}^P$. We shall require that these moves obey similar tiered constraints to those of Eq. \ref{eq:constraints}:
\begin{equation*}
\forall k \leq m \; \exists i, j \leq n, l \in \hat{L}\backslash\{T,B\} \; \text{s.t.} 
\end{equation*}
\begin{eqnarray}\label{eq:constraints2}
k_p = k \Rightarrow ( ( r_p < i \Rightarrow t_p = T ) \wedge ( r_p \geq j \Rightarrow t_p = B ) \nonumber \\
\wedge ( i \leq r_p < j \Rightarrow t_p = l) )
\end{eqnarray}
Our move space $\mathcal{T}$ is thus the subset of $\hat{L}^P$ satisfying constraints in Eq. \ref{eq:constraints2}. Given a move $\mathbf{t}$, we can thus write $i_k$ and $j_k$ for the break points on column $k$, and $l_k \in L$ for the single label taken by $t_p$ for $p$ on column $k$ and rows $i...j-1$.  If $i=j$ we can arbitrarily set $l_k=0$. Given a move $\mathbf{t}$, the transformation function for our move then switches all pixels $p$ for which $t_p \neq \{T, B\}$ to the value $t_p$, and leaves all other variables unchanged. Writing $\mathbb{T}_p(f,\mathbf{t})$ for the value of $\mathbb{T}(f,\mathbf{t})$ at pixel $p$, we can thus express the transformation as:
\begin{eqnarray}\label{eq:transf}
\mathbb{T}_p(f,\mathbf{t}) = \left\{\begin{array}{cc} t_p & \text{if $t_p \neq \{T, B\}$}  \\ f_p & \mathrm{otherwise} \end{array} \right\}
\end{eqnarray}

As described above for the general case, we generate a sequence of labelings $f^{\nu=1...I}$, which is initialized randomly, and satisfies $E(f^1)>E(f^2)...>E(f^I)$.  Using the tiered labeling algorithm of Sec. \ref{sec:tiered}, we can perform an optimal search at each iteration, so that $E(f^{\nu+1}) = \arg\min_{\mathbf{t}\in\mathcal{T}} E(\mathbb{T}(f^{\nu},\mathbf{t}))$.  We achieve this by transforming the search for the optimal $\mathbf{t}$ into a tiered labeling problem as follows.  We can reexpress the energy of a particular transformation $E(\mathbb{T}(f^{\nu},\mathbf{t}))$ directly as an energy over the 
transformation vectors $\mathbf{t}$ as:
\begin{eqnarray}\label{eq:mrfprime}
E^\nu(\mathbf{t}) = \sum_{p\in P}D_p^\nu(t_p) + \sum_{pq\in N} V_{pq}^\nu(t_p, t_q)
\end{eqnarray}
where the data and pairwise terms in Eq. \ref{eq:mrfprime} are defined as:
\begin{eqnarray}
D_p^\nu(t_p) &=& \left\{\begin{array}{cc} D_p(t_p) & \text{if $t_p \in L$}  \\
D_p(f_p^\nu) & \mathrm{otherwise} \end{array} \right\} \nonumber
\end{eqnarray}
\begin{eqnarray}\label{eq:dvprime}
& V_{pq}^\nu(t_p,t_q) =
& \left\{\begin{array}{cc} V_p(t_p,t_q) & \text{$t_p,t_q \in L$}  \\
V_p(t_p,f_q^{\nu}) & \text{$t_p\in L, t_q \in \{T,B\}$}  \\
V_p(f_p^{\nu},t_q) & \text{$t_p\in \{T,B\}, t_q \in L$}  \\
V_p(f_p^{\nu},f_q^{\nu}) & \mathrm{otherwise} \end{array} \right\} \nonumber \\
\end{eqnarray}
It is clear then that solving the minimization:
\begin{eqnarray}
f^{\nu+1} = \arg\min_{\mathbf{t}\in\mathcal{T}} E(\mathbb{T}(f^{\nu},\mathbf{t}))
\end{eqnarray}
is equivalent to minimizing Eq. \ref{eq:mrfprime}, subject to the constraints Eq. \ref{eq:constraints2}, which can thus be achieved via the tiered labeling approach of Sec. \ref{sec:tiered}.  Having found $\mathbf{t}^* = \arg\min_{\mathbf{t}}E^\nu(\mathbf{t})$, we can then generate $f^{\nu+1}$ via:
\begin{eqnarray}\label{eq:gstar}
f^{\nu+1}_p &=& \left\{\begin{array}{cc} t^*_p & \text{if $t^*_p \in L$}  \\
f^{\nu}_p & \mathrm{otherwise} \end{array} \right\}
\end{eqnarray}

As described above, we terminate when $E(f^{\nu+1}) = E(f^{\nu})$. In the discussion above we enforce the tiered constraints on the move space such that the tiered move is applied only in one (vertical) direction. However in general we can apply tiered moves in any direction. For example, we can write $i_k$ and $j_k$ for the break points on the row $k$, and $l_k\in L$ for the single label taken by $t_p$ for $p$ on row $k$, and columns $i...j-1$. Thus, a tiered move along horizontal direction follows these constraints: 
\begin{equation*}
\forall r \leq n \; \exists i, j \leq m, l \in \hat{L}\backslash\{T,B\} \; \text{s.t.}
\end{equation*}
\begin{eqnarray}\label{eq:constraints2}
r_p = r \Rightarrow ( ( k_p < i \Rightarrow t_p = T ) \wedge ( k_p \geq j \Rightarrow t_p = B ) \wedge \nonumber \\
 ( i \leq k_p < j \Rightarrow t_p = l) )
\end{eqnarray}

The algorithm is summarized in Alg. 1.
{\SetAlFnt{\footnotesize}
\begin{algorithm}[t]
    \SetAlgoLined
    \SetKwData{Left}{left}\SetKwData{This}{this}\SetKwData{Up}{up}
    \SetKwFunction{Union}{Union}\SetKwFunction{FindCompress}{FindCompress}
    \SetKwInOut{Input}{input}\SetKwInOut{Output}{output}

    \Input{Energy function $E$, initial labeling $f^1$}
    {\em converged := 0, $\nu$ := 1}\;
     \While{$converged = 0$}
    {
    \SetAlFnt{sf}    Calculate $D^\nu$ and $V^\nu$ for $E^\nu(\mathbf{t})$ (Eq. \ref{eq:dvprime})\;
        Calculate $U$ and $H$ terms (Eq. \ref{eq:dp1})\;
        Solve tiered labeling problem for $\mathbf{t}^* = \arg\min_{\mathbf{t}}E^\nu(\mathbf{t})$\;
        Construct $f^{\nu+1}$ from $\mathbf{t}^*$ (Eq. \ref{eq:gstar})\;
        \uIf{$E(f^{\nu+1}) < E(f^{\nu})$}
        {
           $\nu := \nu+1$\;
        }
        \Else
        {
            {\em converged = 1};
        }
    }
    Return $f^{\nu}$\;
    \caption{{\footnotesize Tiered move making algorithm}}
\end{algorithm}
}

\section{Algorithm properties and analysis}\label{sec:bound}
Since our tiered move making algorithm is based on dynamic programming, it can be applied to arbitrary energy pairwise functions, and is guaranteed to converge, since $E(f^\nu)>E(f^\nu+1)$ at each iteration, and given a finite label set $L$ there are only finitely many labelings. If we make further assumptions about the energy, we can also give certain guarantees on the strength of the local optimum our algorithm achieves, as described in Sec. \ref{sec:opt}.  Further, we discuss the complexity of the algorithm in Sec. \ref{sec:complex}.

\subsection{Optimality}\label{sec:opt}
We show here that we can give guarantees about the local optimum achieved by our algorithm if we restrict the class of energy functions we consider to ones in which $\forall p,l \; D_p(l) \geq 0$, and $\forall p,q,l_1,l_2 \; V_{pq}(l_1,l_2) \geq 0$, with equality occurring if and only if $l_1 = l_2$.  We note that the assumption that the unary terms are non-negative is not restrictive, as we can always add a constant $a$ to all outputs of $D_p(.)$ to create a new potential $D^\prime_p(l)=D_p(l)+a$, and replacing $D_p(.)$ by $D^\prime_p(.)$ simply adds a constant to the global energy $E^\prime(f)=E(f)+a$ which leaves the CRF distribution unchanged. The restriction on the pairwise terms covers many cases of interest, such as semi-metric potentials (which additionally require 
$\forall p,q,l_1,l_2 \; V_{pq}(l_1,l_2) = V_{pq}(l_2,l_1)$), metric potentials (which add to the semimetric requirements $\forall p,q,l_1,l_2,l_3 \; V_{pq}(l_1,l_2) + V_{pq}(l_2,l_3) \geq V_{pq}(l_1,l_3)$), and pairwise Potts potentials (which require $\forall p,q \; \exists v \; \forall l_1\neq l_2 \; V_{pq}(l_1,l_2) = v$).  Additionally, we are interested in labelings of the image that are {\em tiered consistent}; that is, given a labeling $f$, and an associated set of {\em maximal contiguous regions} $R^f$ whose members must take a single label, be connected, and have non-matching neighboring pairs on their boundary (i.e. $R^f = \{\rho \subset P|(\exists l\in L, \; \forall p\in\rho \; (f_p=l)) \; \text{and} \; (\forall p,q \in \rho, \; \exists Z<P,\;(p_1,p_2,...p_Z) \in \rho^Z \; (p=p_1 \wedge q=p_Z \wedge (\forall z<Z \; ((p_z,p_{z+1})\in N)))) \;  \text{and} \; (\forall p,q \in P \; ((p,q)\in N \wedge p\in\rho \wedge q \not\in \rho \Rightarrow f_p \neq f_q))\}$), we can define the set of tiered consistent labelings $C$ as:
\begin{eqnarray}
C &=& \{f|\forall \rho \in R^f,k \leq n \; \exists i,j \; s.t. \nonumber \\
&& k_p = n \wedge (i\leq r_p < j) \Rightarrow p \in \rho\}
\end{eqnarray}
Fig. \ref{fig:tieredConsistent} illustrates the properties introduced above.  Given these definitions, we can make the following guarantee:

\begin{figure}[t]
\begin{center}
  \begin{tabular}{cc}
     \includegraphics[width=0.65\columnwidth]{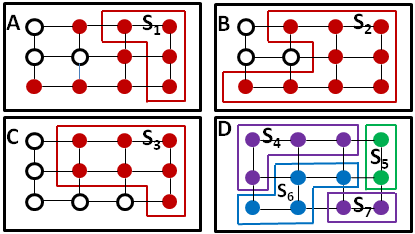}
  \end{tabular}
\end{center}
\caption{\textit{Maximal contiguous regions and tiered consistency.  Colours correspond to different labels. (A) $S_1$ is not a {\em maximally contiguous region}, since there are neighbour pairs on its boundary sharing a common label. (B) $S_2$ is maximally contiguous but not {\em tiered consistent}, since its intersection with column 2 is disconnected (requiring 4 break-points to represent). (C) $S_3$ is both maximally contiguous and tiered consistent. (D) In a {\em tiered consistent labeling} ($f \in C$) all maximally contiguous regions ($S_4, S_5, S_6, S_7 \in R^f$) are tiered consistent. Note that a single label may give rise to multiple maximally contiguous regions, e.g. $S_4$ and $S_7$.  See Sec. \ref{sec:opt}.}}
\label{fig:tieredConsistent}
\end{figure}

\textbf{Theorem 1.} \textit{Let $\hat{f}$ be a local minimum for tiered labeling moves and $f^\dagger = \arg\min_{f\in C}E(f)$ be the optimal tiered consistent labeling. 
Then for energies with non-negative unary terms and in which $\forall p,q,l_1,l_2 \; V_{pq}(l_1,l_2) \geq 0$, with equality occurring if and only if $l_1 = l_2$, we have 
that $E(\hat{f}) \leq 2 \kappa E(f^\dagger)$, where $\kappa=\max_{p,q \in N}(\max_{l_1\neq l_2 \in L}V_{pq}(l_1,l_2)/ \min_{l_1\neq l_2 \in L}V_{pq}(l_1,l_2))$.}
\textit{Proof: }
We consider the set of maximally contiguous regions $\rho \in R^{f^\dagger}$.  Given any $\rho$, we define $f^\rho$ to be:
\begin{eqnarray}
f^\rho = \left\{\begin{array}{cc} f^\dagger_p & \text{if $p \in \rho$}  \\ \hat{f}_p & \mathrm{otherwise} \end{array} \right\}
\end{eqnarray}
We can assume $E(\hat{f})\geq E(f^\dagger)$, since otherwise the bound follows directly. Since $f^\rho$ is within a tiered move of $\hat{f}$, we have:
\begin{eqnarray}\label{eq:ft}
E(f^\dagger) \leq E(\hat{f}) \leq E(f^\rho)
\end{eqnarray}
Further, we can define the sets $I^\rho = \rho\cup\{(p,q)\in N| p,q \in \rho\}$, $B^\rho = \{(p,q)\in N| p \in \rho, q \not\in \rho\}$, and $O^\rho = (P\backslash \rho)\cup\{(p,q)\in N| p,q \not\in \rho\}$, which represent respectively, all pixels and neighbour pairs falling inside $\rho$ ($I^\rho$), neighbour pairs between $\rho$ and $P\backslash \rho$ ($B^\rho$), and all pixels and neighbours outside of $\rho$($O^\rho$).  We then have:
\begin{eqnarray}\label{eq:OIB}
E(f^\rho|O^\rho) &=& E(\hat{f}|O^\rho) \nonumber \\
E(f^\rho|I^\rho) &=& E(f^\dagger|I^\rho) \nonumber \\
E(f^\rho|B^\rho) &\leq& \kappa E(f^\dagger|B^\rho)
\end{eqnarray}
where we write $E(f|O^\rho)$ for the value of the energy Eq. \ref{eq:mrf} restricted to potentials on pixels/neighbours in $O^\rho$, and similarly for $E(f|I^\rho)$, $E(f|B^\rho)$. The constant $\kappa$ is defined as stated in the Theorem, $\kappa=\max_{p,q \in N}(\max_{l_1\neq l_2 \in L}V_{pq}(l_1,l_2)/ \min_{l_1\neq l_2 \in L}V_{pq}(l_1,l_2))$, and we can guarantee the last of these inequalities, since the requirement that $\rho$ is a {\em maximal} contiguous set implies that $\forall (p,q)\in B^\rho \; f_p^\dagger \neq f_q^\dagger$, and the requirements on the energy imply therefore $\forall (p,q)\in B^\rho \; V_{pq}(f^\dagger_p,f^\dagger_q)>0$.
Now, by separating the final inequality in Eq. \ref{eq:ft}, we get the expansion $E(\hat{f}|I^\rho) + E(\hat{f}|B^\rho) + E(\hat{f}|O^\rho) \leq E(f^\rho|I^\rho) + E(f^\rho|B^\rho) + E(f^\rho|O^\rho)$, and substituting in the values from Eq. \ref{eq:OIB} we get $E(\hat{f}|I^\rho) + E(\hat{f}|B^\rho) + E(\hat{f}|O^\rho) \leq E(f^\dagger|I^\rho) + \kappa E(f^\dagger|B^\rho) + E(\hat{f}|O^\rho)$, which reduces to $E(\hat{f}|I^\rho) + E(\hat{f}|B^\rho) \leq E(f^\dagger|I^\rho) + \kappa E(f^\dagger|B^\rho)$.
To bound the whole energy, we must sum this final inequality over all $\rho \in R^{f^\dagger}$, giving $\sum_{\rho \in R^{f^\dagger}}(E(\hat{f}|I^\rho) + E(\hat{f}|B^\rho)) \leq \sum_{\rho \in R^{f^\dagger}}(E(f^\dagger|I^\rho) + \kappa E(f^\dagger|B^\rho))$.  Letting $B=\bigcup_{\rho \in R^{f^\dagger}}B^\rho$ and collecting together terms, we can reexpress this as:
\begin{eqnarray}\label{eq:bound1}
E(\hat{f})+E(\hat{f}|B) \leq E(f^\dagger) + (2\kappa - 1) E(f^\dagger|B)
\end{eqnarray}
and subtracting $E(\hat{f}|B)\geq0$ from the left hand side of Eq. \ref{eq:bound1} and adding $(2\kappa - 1)(E(f^\dagger)-E(f^\dagger|B))\geq0$ to the right we have the final bound $E(\hat{f}) \leq 2 \kappa E(f^\dagger)$. $\;\;\;\;\;\;\;\;\;\;\;\;\;\;\;\;\;\;\;\;\;\;\;\;\;\;\;\;\;\;\;\;\;\;\;\;\;\;\;\;\;\square$

\textbf{Corollary.} \textit{For energies of the form in Theorem 1, and for which the optimal labeling $f^*=\arg\min_f E(f)$ is tiered consistent, we have that $f^* = f^\dagger$, and thus we achieve the global bound $E(\hat{f}) \leq 2 \kappa E(f^*)$.}

For cases in which the optimal energy $f^*$ is tiered consistent, we can thus achieve the same bound as $\alpha$-expansion \cite{Boykov_pami01} for energies with metric and Potts pairwise potentials.  We note that, when $f^\dagger = f^*$, this bound actually holds for energies with a larger class of pairwise potentials as specified by Theorem 1, including semi-metrics and a restricted class of non-symmetric potentials.  In fact, for the pairwise case these are the same restrictions on the energy as in a recent generalization of the $\alpha$-expansion bound in \cite{gouldAlphabetSoup}.  Our bound is thus identical to \cite{gouldAlphabetSoup} for energies with a tiered consistent $f^*$, although we cannot apply the \cite{gouldAlphabetSoup} bound directly to the analyze tiered moves, since they do not form an optimal $\gamma$-expansion move in the sense required by \cite{gouldAlphabetSoup}.

Clearly, we cannot know for any given energy whether or not $f^\dagger = f^*$.  For many problems though, such as segmentation of objects with approximately convex shapes, or stereo, where we can expect regions of constant disparity to be approximately vertical, we can expect that this condition will hold approximately, or equivalently that the violations of the tiered constraints in $f^*$ will be small (assuming the energy itself provides a good representation of the problem).  This argument lends theoretical support to our claim in Sec. \ref{sec:exp} that tiered moves are a competitive alternative to $\alpha$-expansion moves (and non-metric variants) for a wide range of problems.

\begin{figure}[t]
\begin{center}
  \begin{tabular}{cc}
     \includegraphics[width=0.60\linewidth]{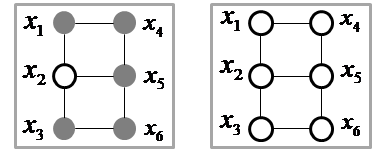}
  \end{tabular}
\end{center}
\caption{\textit{Suboptimal binary case: Left shows the optimal configuration $f^*$ which is non-tiered consistent ($E(f^*)=1$), but right shows the best tiered consistent labeling $f^\dagger$ having suboptimal energy ($E(f^\dagger)=Q$). White indicates variables with value 0, and black indicates values 1. See Sec. \ref{sec:opt} (Worst case suboptimality).}}
\label{fig:suboptimal_case}
\end{figure}

\textit{Worst case suboptimality:}  Although we have argued above that for many problems of interest, we might expect $f^\dagger \approx f^*$, we now construct a simple binary example to show that in general the energy of $f^\dagger$ can be arbitrarily far from the optimum.  Consider for example a binary energy over $P=6$ pixels with $m=3$ rows and $n=2$ columns, with column-major indexing (see Fig. \ref{fig:suboptimal_case}).  Assuming a 4-connected graph, and Potts pairwise terms between neighbours, we define an energy such that $D_{1}(0)=D_{3}(0)=Q/2$ and $D_{2}(1)=2Q$ ($Q>1$), while the non-zero term for pairwise potentials $V_{1,2}$,$V_{2,3}$ and $V_{2,5}$ is set to $1/3$ and for $V_{1,4}$,$V_{4,5}$,$V_{5,6}$ and $V_{3,6}$ is set to $2Q$.  Writing $f = \{x_{1},x_{2},x_{3},x_{4},x_{5},x_{6}\}$ for a joint setting of the 6 variables, we can determine that the global optimum is $f^* = \{1,0,1,1,1,1\}$, where $E(f^*)=1$.  This labeling however is not tiered consistent, as it contains a maximal contiguous region of 1's with 4 break-points on the first column.  The optimal tiered consistent configuration is $f^\dagger = \{0,0,0,0,0,0\}$, where $E(f^\dagger)=Q$ (see Fig. \ref{fig:suboptimal_case}). For all other labelings $E(f) \geq 2Q$.  As well as being the optimal tiered consistent labeling, $f^\dagger$ is also a local optimum for tiered moves (i.e. it forms a possible $\hat{f}$), since the only configuration with lower energy cannot be reached in a single move (in general, the algorithm can converge to arbitrary labelings). We note that, by construction $E(f^\dagger)=Q$ can be increased without limit, and so can be arbitrarily far from the global optimum $E(f^*)=1$.  This example shows that our algorithm can fail when the global optimum contains closely connected non-convex regions.

\subsection{Complexity}\label{sec:complex}

The complexity of the individual moves in our algorithm can be derived from the complexity of the tiered labeling algorithm itself, which dominates the move:

\textbf{Theorem 2.} \textit{The time complexity of each individual tiered move is $O(nm^2K^2)$, where $n,m$ are the number of image columns and rows respectively, and $K$ 
is the number of labels in the global energy.}

\textit{Proof:}
We first note that application of the tiered labeling algorithm dominates the move, since computing $D^\nu$, $V^\nu$, $U$ and $H$ are all $O(nmK)$, and applying Eq. \ref{eq:gstar} to update the solution is $O(mn)$.  As described in Sec. \ref{sec:move}, finding the optimal move involves solving a tiered labeling problem with an expanded label set $\hat{L}=L\cup\{T,B\}$.  Since labels $T,B$ cannot appear in the dynamic programming variables $s_{k=1...n}=(i_k,j_k,l_k)$ we have a complexity $O(nm^2K^2)$ for each move, following \cite{tiered_cvpr_felzenszwalbV10}.$\;\;\;\;\;\;\;\;\;\;\;\;\;\;\;\;\;\;\;\;\;\;\;\;\;\;
\;\;\;\;\;\;\;\;\;\;\;\;\;\;\;\;\;\;\;\;\;\;\;\;\;\;\;\;\;\;\;\;\;\;\;\;\;\;\;\;\;\;\;\;\;\;\;\;\;\;\;\;\;\;\;\;\;\;\;\;\;\;\square$

In practice, on the tsukuba image (Fig. ~\ref{fig:stereo_images}) of size 384$\times$288 with 16 disparity labels for the stereo problem, our method takes almost 5 minutes, compared to $\alpha$-expansion which takes almost 1 sec, and TRWS which takes 6 secs. Each tiered move iteration takes 18 secs, and the algorithm takes almost 15 moves to converge. In general we reach an energy value very close to the optimal solution within few iterations. Further, as noted in Felzenswalb et.al. [2], the tiered labelling method can be trivially implemented on a GPU. One possible GPU implementation would involve distributing the computation per state over threads. A minimization must be performed per thread across the states of the previous site (column), and an outer minimization also made across the states of the final site. These min-calculations across states would be the computational bottleneck which could be achieved in $O(\log(n^2K))$ time on a GPU. The calculations per site would have to be done sequentially, since each site is dependent on the results of the previous site. We believe this implementation would be highly efficient compared to graph-cuts and TRWS on a GPU. Thus an optimized GPU implementation would substantially reduce the timing and make our algorithm a competitive choice for many real problems.


\section{Experiments}\label{sec:exp}

We demonstrate the performance of our inference methods on two broad problem domains. In the first case, our pairwise terms take an unconstrained general form resulting from learning an arbitrary function on the training dataset. Our second experimental set-up restricts the pairwise terms to take certain basic models like Potts, (truncated) linear, or (truncated) quadratic. Details of all the experiments are provided in the following subsections.

\subsection{Learnt Pairwise Terms}

We outline here experimental results on the PascalVOC-11 segmentation dataset~\cite{everingham_voc_ijcv08} for the object class segmentation problem. This dataset consists of 1112 training images, 1111 validation images, and 20 object classes. The task is to assign every pixel in the image an object label such as car, person, etc. or background. Our energy function includes unary and pairwise terms, where the unary potentials are based on dense responses from a Textonboost style classifier~\cite{shotton_texboost_eccv06}, trained by boosting classifiers defined on multiple dense features defined on colour, textons, histogram of orientated gradients (HOG), SIFT and pixel location. 

\begin{figure}
\begin{center}
  \begin{tabular}{cc}
      \includegraphics[width=0.15\linewidth]{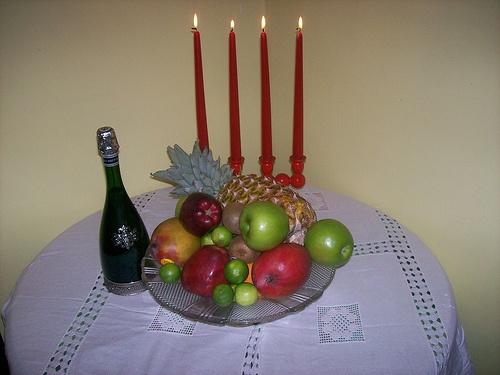}
      \includegraphics[width=0.15\linewidth]{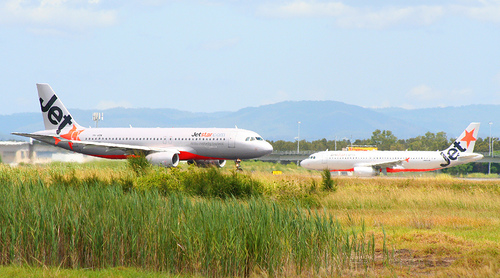}
      \includegraphics[width=0.15\linewidth]{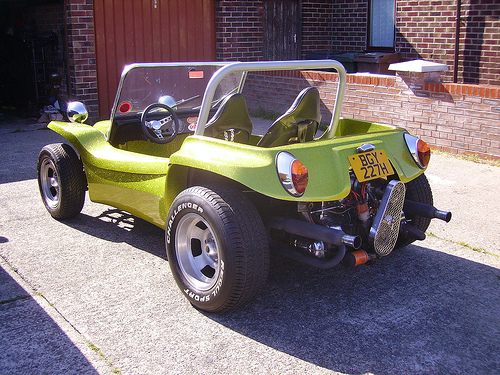}
      \includegraphics[width=0.15\linewidth]{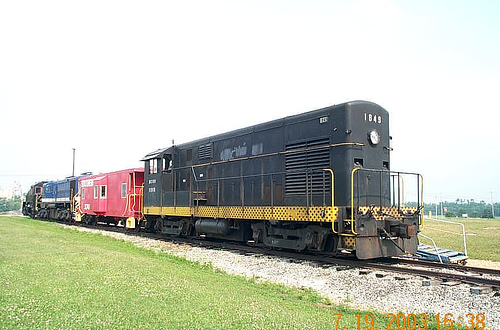}
      \includegraphics[width=0.15\linewidth]{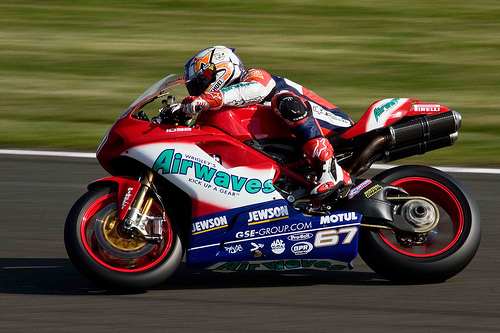}
      \includegraphics[width=0.15\linewidth]{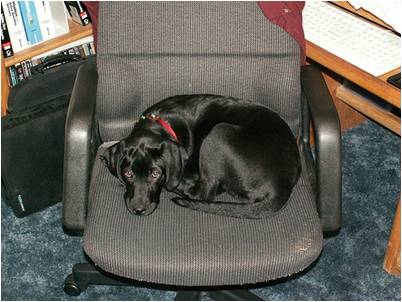} \\
      \includegraphics[width=0.15\linewidth]{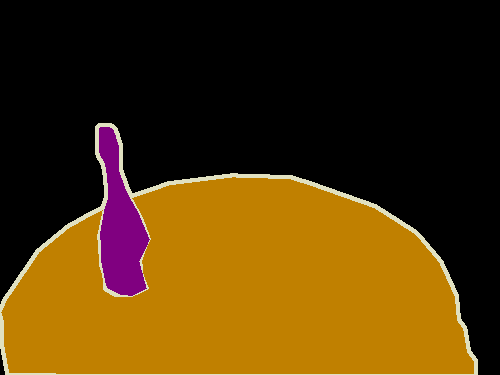}
      \includegraphics[width=0.15\linewidth]{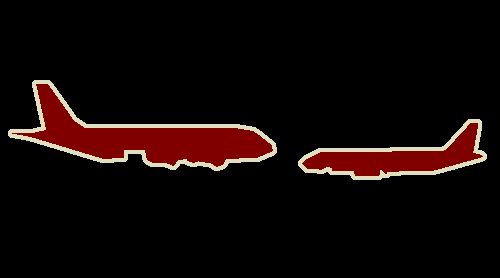}
      \includegraphics[width=0.15\linewidth]{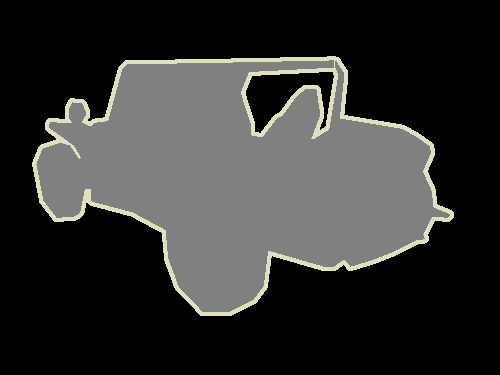}
      \includegraphics[width=0.15\linewidth]{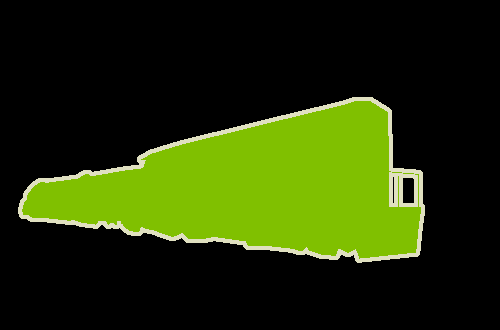}
      \includegraphics[width=0.15\linewidth]{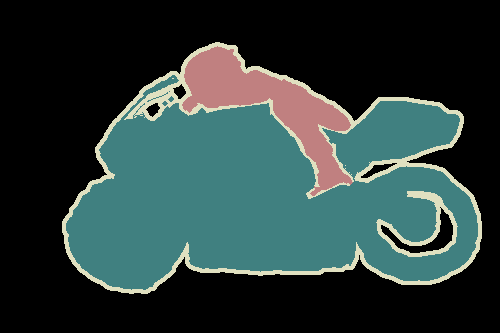}
      \includegraphics[width=0.15\linewidth]{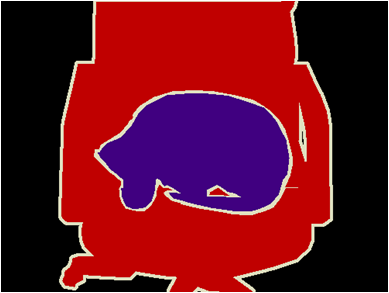} \\
      \includegraphics[width=0.15\linewidth]{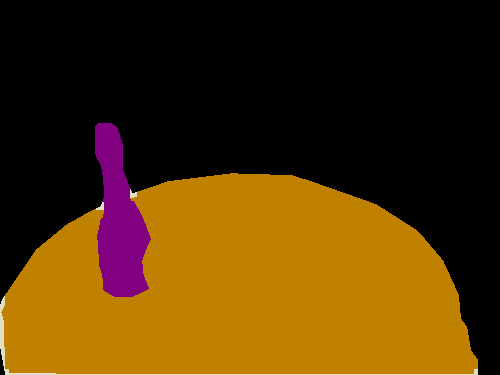}
      \includegraphics[width=0.15\linewidth]{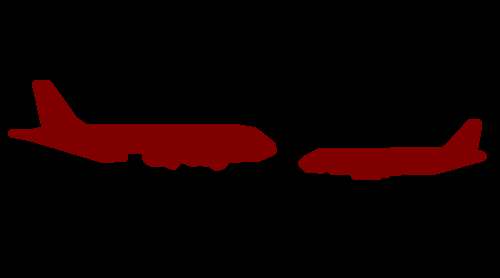}
      \includegraphics[width=0.15\linewidth]{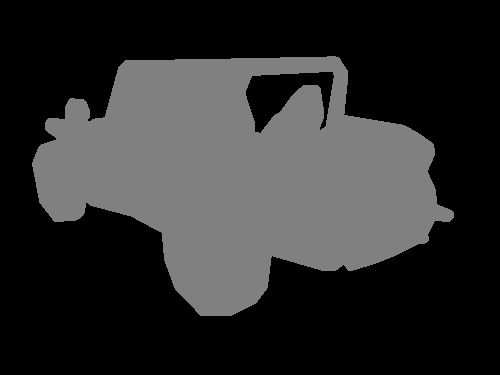}
      \includegraphics[width=0.15\linewidth]{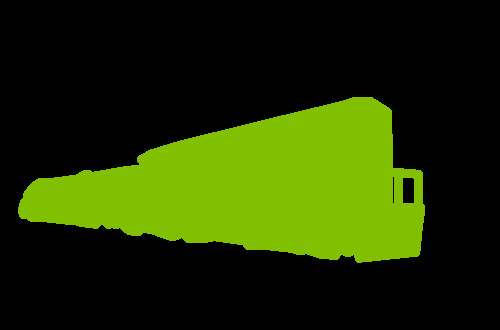}
      \includegraphics[width=0.15\linewidth]{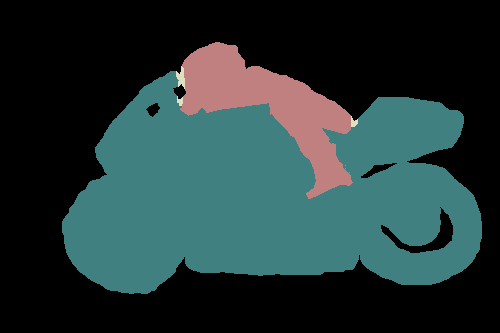}
      \includegraphics[width=0.15\linewidth]{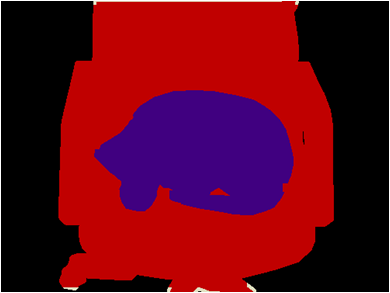} \\
  \end{tabular}
\end{center}
\vspace{-3mm}
\caption{\textit{Shown are some of the images from our boundary dataset. First row: original images from PascalVOC-11 segmentation dataset. Second row: ground-truth images with void pixels along the interclass boundaries. Third row: images from our boundary dataset. We use these images for learning class-class and class-background pairwise interaction terms for the object class segmentation problem.}}
\label{fig:boundary_images}
\end{figure}

We formulate the problem of learning the general class-class and class-background pairwise terms in a max-margin framework [see \cite{tsochan_icml04}] under a supervised setting. Given the input and output pairs for the $i^{th}$ image $(x_i, f_i)$ where $i\in I$, the max-margin learning framework learns the parameters of the model by maximizing a scoring function $S_w(\tilde{f}) = \mathrm{argmax}_{\tilde{f}\in F}<w,\phi(x_i, \tilde{f})>$, by minimizing a loss function between the true output $f_i$ and the predicted output $\tilde{f}_i$. Here weight vector $w$ is the learnt pairwise terms. We follow the method of Tsochantaridis et.al.~\cite{tsochan_icml04} who use a hamming distance based loss function. Our class-class or class-background features are based only on colour differences between pair of pixels. Thus, in order to generate the feature vector, we create a boundary dataset where we augment and update the PascalVOC-11 segmentation ground-truth dataset~\cite{everingham_voc_ijcv08} by completely annotating the boundary regions. A set of ground-truth images from our boundary dataset is shown in the Figure~\ref{fig:boundary_images} along with their corresponding original and Pascal ground-truth images. To train the pairwise terms, we split the images into neighbouring 2 pixel sub-images, and use a sample of these as our training examples. We access overall percentage of pixels correctly classified, the average recall and intersection/union (I/U) score per class defined in terms of the true/false positives/negatives for a given class as TP/(TP+FP+FN)). We first compare our model with a graph-cuts based $\alpha$-expansion method using the Potts model, where the parameters of the Potts model are set by cross validation. We observe both $\alpha$-expansion and our $t$-move methods achieve the same $23.91\%$ I/U score. We observe a slight improvement in I/U accuracy by using the learnt pairwise terms to $24.03\%$. Table~\ref{tab:pascal_results} gives details of the results. 

\begin{table}
\begin{center}
    \begin{tabular}{|l|l|l|l|l|} \hline
     Method & Pairwise Terms & Overall & Av. Recall & I/U \\ \hline
     $\alpha$-expansion & Potts & 71.99 & 38.96 & 23.91 \\
     t-move & Potts & 71.99 & \textbf{38.96} & 23.91 \\
     t-move & Learnt & \textbf{72.81} & 37.92 & \textbf{24.03} \\ \hline
    \end{tabular}
\end{center}
\vspace{-3mm}
\caption{\textit{The results provide a quantitative comparison between the learnt pairwise terms and the Potts model. We observe that using the learnt pairwise terms gives a slight improvement in the overall accuracy.}}
\label{tab:pascal_results}
\end{table}

\subsection{Fixed-form Pairwise Terms}

\begin{figure}
\begin{center}
  \begin{tabular}{cc}
     \includegraphics[width=0.15\linewidth]{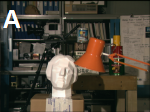}
     \includegraphics[width=0.15\linewidth]{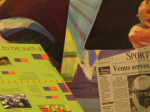}
     \includegraphics[width=0.15\linewidth]{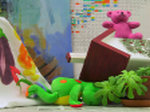}
     \includegraphics[width=0.15\linewidth]{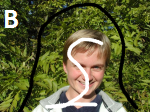}
     \includegraphics[width=0.15\linewidth]{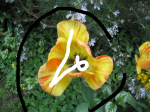}
     \includegraphics[width=0.15\linewidth]{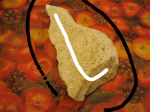} \\
     \includegraphics[width=0.15\linewidth]{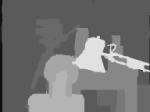}
     \includegraphics[width=0.15\linewidth]{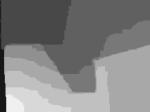}
     \includegraphics[width=0.15\linewidth]{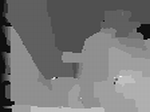}
     \includegraphics[width=0.15\linewidth]{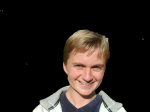}
     \includegraphics[width=0.15\linewidth]{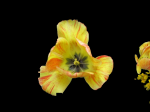}
     \includegraphics[width=0.15\linewidth]{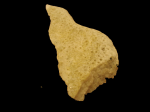} \\
     \includegraphics[width=0.23\linewidth]{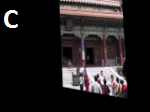}
     \includegraphics[width=0.23\linewidth]{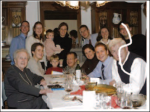}
     \includegraphics[width=0.23\linewidth]{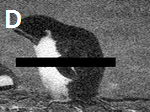}
     \includegraphics[width=0.23\linewidth]{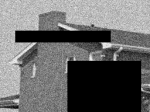} \\
     \includegraphics[width=0.23\linewidth]{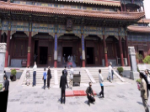}
     \includegraphics[width=0.23\linewidth]{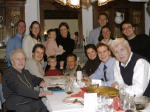}
     \includegraphics[width=0.23\linewidth]{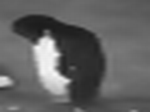}
     \includegraphics[width=0.23\linewidth]{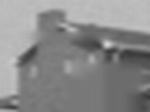}
  \end{tabular}
\end{center}
\vspace{-3mm}
\caption{\textit{Shown are the input (first row) and output (second row) images of our tiered move making algorithm corresponding to different benchmark problems. (From left to right) \textbf{Stereo (A)}: tsukuba (192$\times$144, 8 disparities), venus (217$\times$192, 10 disparities) and teddy (112$\times$93, 15 disparities) images; \textbf{Segmentation (B)}: person (600$\times$450), flower (600$\times$450) and sponge (640$\times$480) images; \textbf{Image stitching (C)}: pano (178$\times$120, 7 labels), and family (376$\times$283, 5 labels); \textbf{Image denoising (D)}: penguin (30$\times$44, 256 labels), and family (32$\times$32, 256 labels).}}
\label{fig:benchmark_images}
\end{figure}

Further we present results on several low-level benchmark problems, including stereo correspondence, image segmentation, image stitching and image denoising. The review paper of Szeliski et. al.~\cite{Szeliski_pami08} provides some benchmark images corresponding to each problem, some of which we rescale (see Fig.~\ref{fig:benchmark_images}). We have two versions of our tiered move method: $t$-move1 and $t$-move2. In $t$-move1 tiered labeling is performed only along rows, however in $t$-move2 tiered labeling is performed along both rows and columns. We show the results of $t$-move2 when the results are significantly different from the results of $t$-move1. We compare our methods with four other energy minimization methods, $\alpha$-expansion, max-product loopy belief propagation (LBP), and sequential tree re-weighted message passing (TRWS) on these benchmark applications, and also the original tiered labeling algorithm \cite{tiered_cvpr_felzenszwalbV10} for segmentation. We use code provided by the authors for the QPBO method~\cite{kolmogorov_qpbo} and codes in the software library provided by Szeliski et. al.~\cite{Szeliski_pami08} for LBP, $\alpha$-expansion and TRWS. We use QPBO to perform expansion moves over a series of iterations similar to $\alpha$-expansion, as in Woodford et.al.~\cite{woodford2006fields}. The pairwise potentials we use for our comparative study are Potts models, truncated/non-truncated linear and quadratic functions. In all these examples, we use the same unary potential as used by~\cite{Szeliski_pami08}. 

\textit{Stereo:} Given left and right rectified images of a scene, the stereo problem requires us to calculate the disparity of each pixel in the left image to a pixel in  right image.  Discrete labels are used to represent disparities in terms of pixel displacement.  We conduct experiments on the tsukuba, venus, and teddy images; these  three benchmark images are shown in Fig. ~\ref{fig:benchmark_images}(\textbf{A}). 

Table~\ref{tab:stereo_results} shows the comparative lowest energy values achieved by $\alpha$-expansion, LBP, TRWS and tiered moves ($t$-move1, $t$-move2) on the three benchmark images. The smoothness terms are a Potts model, and linear and quadratic pairwise potentials. We observe that our methods consistently achieve lower energy values than $\alpha$-expansion, QPBO, and LBP on all three images. Our $t$-move2 achieves lower energy values than TRWS on the tsukuba and venus images for the quadratic model. We also observe that our $t$-move2 consistently achieves lower energies than $t$-move1. Further, the lowest energy values achieved by our methods are almost equal to the lower bound on the energy values provided by TRWS method. In all these cases, $t$-move2 is able to achieve almost $1\%-2\%$ lower energies than $\alpha$-expansion, QPBO and LBP methods, though TRWS achieves slightly better energies than our method in few cases. 

\begin{table}
\begin{center}
    \begin{tabular}{|l|l|l|l|l|} \hline
    Image & Method & Potts & L1-norm & L2-norm \\ \hline
    \multirow{6}{*}{Tsukuba} & Lower Bound & 84730 & 106407 & 119103 \\
    & LBP & 90930 & 112659 & 124668 \\
    & TRWS & 84730 & 106407 & 119514 \\
    &$\alpha$-exp & 84870 & 106529 & 120880 \\
    & QPBO & 90247 & 113593 & 128212 \\
    & $t$-move1 & 84745 & 106411 & 120003 \\
    & $t$-move2 & 84745 & 106407 & 119280 \\ \hline
    \multirow{6}{*}{Venus} & Lower Bound& 142403 & 160417 & 137608 \\
    & LBP & 147133 & 160417 & 139808 \\
    & TRWS & 142403 & 160417 & 138726 \\
    &$\alpha$-exp & 142549 & 160512 & 139286 \\
    & QPBO & 156008 & 175850 & 146927 \\
    & $t$-move1 & 142447 & 160436 & 138844 \\ 
    & $t$-move2 & 142424 & 160432 & 137966 \\ \hline
    \multirow{6}{*}{Teddy} & Lower Bound & 115479 & 126599 & 139117 \\
    & LBP & 117018 & 128843 & 141233 \\
    & TRWS & 115479 & 126599 & 139341 \\
    &$\alpha$-exp & 115829 & 126631 & 139949 \\
    & QPBO & 120220 & 131607 & 145384 \\
    & $t$-move1 & 115479 & 126657 & 139789 \\
    & $t$-move2 & 115479 & 126649 & 139394 \\ \hline
    \end{tabular}
\end{center}
\vspace{-3mm}
\caption{\textit{Stereo: Compares the lowest energy values achieved by tiered moves and other benchmark energy minimization methods on three different stereo images for Potts, linear and quadratic pairwise potentials. The lower bound on the energy values given by TRWS is also shown.}}
\label{tab:stereo_results}
\end{table}

\textit{Segmentation:} We select the person, flower, and sponge images as shown in Fig. ~\ref{fig:benchmark_images}(\textbf{B}) for binary segmentation. Our data-term is based on gaussian mixture colour models of the foreground and background, and the smoothness term is a Potts model modulated by local contrast values. The corresponding energy values for all these images and benchmark algorithms are shown in Table~\ref{tab:segmentation_results}. We achieve the global minimum energy values on all images, as does $\alpha$-expansion. Further, we achieve almost $2\%-3\%$ lower energy values than the original tiered labeling method on all images.

\begin{table}
\begin{center}
    \begin{tabular}{|l|l|l|l|l|} \hline
    Image & Method & Smooth & Data & Total \\ \hline
    \multirow{3}{*}{Person} & $t$-label & 682815 & 32890529 & 33573344 \\
    &$\alpha$-exp & 647809 & 30157043 & 30804852 \\
    &$t$-move1 & 647809 & 30157043 & 30804852 \\ \hline
    \multirow{3}{*}{Flower} & $t$-label & 636004 & 40235734 & 40871738 \\
    &$\alpha$-exp & 679744 & 39831530 & 40511274 \\
    &$t$-move1 & 679744 & 39831530 & 40511274 \\ \hline
    \multirow{3}{*}{Sponge} & $t$-label & 273505 & 31974735 & 32248240 \\
    &$\alpha$-exp & 255933 & 31471232 & 31727165 \\
    &$t$-move1 & 255908 & 31471257 & 31727165 \\ \hline
    \end{tabular}
\end{center}
\vspace{-3mm}
\caption{\textit{Segmentation: We compare the lowest energy values achieved by different benchmark energy minimization methods on three different segmentation images. We achieve the global minimum in all cases, as does $\alpha$-expansion. Results from the original tiered labeling algorithm ($t$-label) \cite{tiered_cvpr_felzenszwalbV10} are also given.}}
\label{tab:segmentation_results}
\end{table}

\textit{Photomontage:} Given a set of overlapping images, photomontage requires us to stitch them into one seamless image. The data-term is either 0 or infinity, depending on whether that pixel is in the field of view of the current camera direction. We show our results on the family and panorama images in Fig. ~\ref{fig:benchmark_images}(\textbf{C}). Our pairwise terms depend on pixel location and labels, as given in Szeliski et.al.~\cite{Szeliski_pami08}. In this case, our method achieves lower energy values than all the other  benchmark methods shown in Table~\ref{tab:pano_results}. We achieve almost $0.2\%-0.5\%$ lower energy than TRWS and $\alpha$-expansion on the family image set. For the pano image, we achieve the lower bound on the energy value (i.e. the global optimum) as achieved by all the other methods.

\begin{table}
\begin{center}
    \begin{tabular}{|l|l|l|l|l|} \hline
    Image & Method & Smooth & Data & Total \\ \hline
    \multirow{5}{*}{(148956)} & TRWS & 149784 & 0 & 149784 \\
    Family & $\alpha$-exp &149181 & 0 & 149181 \\
    & $t$-move1 & 149024 & 0 & 149024 \\ \hline
    \multirow{5}{*}{(80813)} & TRWS & 80813 & 0 & 80813 \\
    Pano & $\alpha$-exp & 80813 & 0 & 80813 \\
    & $t$-move1 & 80813 & 0 & 80813 \\ \hline
    \end{tabular}
\end{center}
\vspace{-3mm}
\caption{\textit{Image Stitching: Compares lowest energy values achieved by different benchmark energy minimization methods on two different image stitching benchmarks. The lower bound on energy values given by TRWS is also shown in the first column.}}
\label{tab:pano_results}
\end{table}

\textit{Image denoising:} In the denoising problem, we are given a noisy image, and the task is to output a denoised image. This problem is also formulated as an energy minimization problem over an MRF where each pixel can receive any label from 256 intensity values. We conduct experiments on the house and penguin images in Fig. ~\ref{fig:benchmark_images}(\textbf{D}). Our data term is the squared difference between the actual pixel value and the output label, and the smoothness term is the squared difference between labels, weighted by a common factor $\lambda = 5$. In all these benchmark images, we achieve energies lower than $\alpha$-expansion by almost $5\%$, although our energies are slightly higher than TRWS (see Table~\ref{tab:denoising_results}).

\begin{table}
\begin{center}
    \begin{tabular}{|l|l|l|l|l|} \hline
    Image & Method & Smooth & Data & Total \\ \hline
    \multirow{5}{*}{(1776156)} & TRWS & 270425 & 1505731 & 1776156 \\
    Penguine & $\alpha$-exp & 210075 & 1778962 & 1989037 \\
    & $t$-move1 & 261150 & 1521812 & 1782962 \\ 
    & $t$-move2 & 259485 & 1518729 & 1778214 \\ \hline
    \multirow{5}{*}{(605613)} & TRWS & 181285 & 424329 & 605614 \\
    House & $\alpha$-exp & 202670 & 419748 & 622418 \\
    & $t$-move1 & 179710 & 426549 & 606259 \\
    & $t$-move2 & 181548 & 424384 & 605932 \\ \hline
    \end{tabular}
\end{center}
\vspace{-3mm}
\caption{\textit{Denoising: Compares lowest energy values achieved by different benchmark energy minimization methods on two different image denoising images. The lower bound on energy values given by TRWS is shown in the first column.}}
\label{tab:denoising_results}
\end{table}


\section{Discussion and Conclusion}\label{sec:disc}

In this paper, we propose a tiered move making algorithm. Each iteration of algorithm makes a decision at each pixel whether to retain the current label or to take a new one based on the optimal tiered move step. Each optimal tiered move is found by applying the dynamic programming based tiered labeling method.

Our method is guaranteed to converge to a local minima for any pairwise potential. Graph cuts based $\alpha$-expansion will not converge to a local minimum when functions are non-metric and non-submodular, since generally $\alpha$-expansion solvers truncate the pairwise potential. Similarly, message passing based methods like LBP and TRWS are not guaranteed to converge and the final solution may oscillate between two different labelings. However, the TRWS method interestingly provides a lower bound on the energy, which provides a basis of comparison for other algorithms. Our tiered move method always converges to a local optimum, and does not suffer from the problem of oscillation.

The QPBO method~\cite{kolmogorov_qpbo} is a graph-cut based method which is able to handle non-submodular energy functions. But, this method does not always label all the nodes, which leads to partial labelings. The unlabeled pixels are labeled heuristically. However, we do not suffer from this problem since we do not allow fractional solutions.

Our tiered move making algorithm however suffers from certain limitations. The time complexity of this method is $O(nm^2K^2)$ per move, and in practice this method is almost 10-15 times slower than $\alpha$-expansion. However, this move complexity could be reduced by using schemes such as $\gamma$-expansions \cite{gouldAlphabetSoup} where only subsets of the labels are expanded at each move, without affecting the guarantees in Sec. \ref{sec:opt}.  Further, as with the tiered labeling method, the space complexity is $O(nm^2K)$ due to the dynamic programming requirements. Also, we cannot guarantee that a series of tiered moves will lead to the globally optimal solution even in the 2 label sub-modular case as is guaranteed by graph cuts (see Sec. \ref{sec:opt}). However, empirically we achieve the global minimum for many of the problems investigated, and we are mostly within $0.001\%$ of the global minimum otherwise. We believe that only if there are many closely connected non-convex shapes in the solution will our method get stuck in bad local minima.

As our experiments suggest, we consistently do well compared to $\alpha$-expansion, LBP, and QPBO and are very competitive with the TRWS method on a wide range of problems, including object class segmentation, stereo correspondence, image stitching, image denoising. Our theoretical analysis provides further support for the expectation that our algorithm will remain competitive in many commonly encountered problems, and the code and dataset is available for download at http://cms.brookes.ac.uk/staff/VibhavVineet/. 



\section*{Acknowledgment}

We thank Paul Sturgess for his valuable help in generating the Pascal boundary dataset. The work was supported by the EPSRC and the IST programme of the European Community, under the PASCAL2 Network of Excellence. Professor Philip H.S. Torr is in receipt of a Royal Society Wolfson Research Merit Award.

\end{document}